\documentclass[letterpaper, 10 pt, conference]{ieeeconf}  

\IEEEoverridecommandlockouts                              

\usepackage[utf8]{inputenc} 
\usepackage[T1]{fontenc}    
\usepackage{url}            
\usepackage{booktabs}       
\usepackage{amsfonts}       
\usepackage{nicefrac}       
\usepackage{microtype}      
\usepackage{xcolor}         
\usepackage{subfig}
\usepackage{graphicx}
\usepackage{algorithm}
\usepackage{algpseudocode}
\usepackage{fancyhdr}
\fancypagestyle{plain}{\pagestyle{fancy}}
\usepackage{etoolbox}

\graphicspath{{images/}}
\newcommand{\approach}{CMZ-DRIL}

\title{\LARGE \bf
Continuous Mean-Zero Disagreement-Regularized Imitation Learning (\approach{})\thanks{This work has been submitted to the IEEE for possible publication. Copyright may be transferred without notice, after which this version may no longer be accessible.}
}

\author{
    Noah Ford$^{1}$, Ryan W. Gardner$^{1}$, Austin Juhl$^{1}$, Nathan Larson$^{1}$
    \thanks{$^{1}$Authors are with Johns Hopkins University Applied Physics Laboratory, Laurel, MD 20723, USA
        {\tt\small \{noah.ford, ryan.gardner, austin.juhl, nathan.larson\}@jhuapl.edu}}
}%

\patchcmd{\maketitle}
  {\end{titlepage}}
  {\thispagestyle{titlepagestyle}\end{titlepage}}
  {}{}

\fancypagestyle{titlepagestyle}
{
  \fancyhf{}
  \fancyfoot[C]{
    NAVAIR Public Release 2023-0156. DISTRIBUTION STATEMENT A. Approved for public release: distribution unlimited.}
  
}

\begin{document}

\maketitle

\begin{abstract}
Machine-learning paradigms such as imitation learning and reinforcement learning can generate highly performant agents in a variety of complex environments. However, commonly used methods require large quantities of data and/or a known reward function. This paper presents a method called Continuous Mean-Zero Disagreement-Regularized Imitation Learning (CMZ-DRIL) that employs a novel reward structure to improve the performance of imitation-learning agents that have access to only a handful of expert demonstrations.
CMZ-DRIL uses reinforcement learning to minimize uncertainty among an ensemble of agents trained to model the expert demonstrations. 
This method does not use any environment-specific rewards, but creates a continuous and mean-zero reward function from the action disagreement of the agent ensemble.  As demonstrated in a waypoint-navigation environment and in two MuJoCo environments, CMZ-DRIL can generate performant agents that behave more similarly to the expert than primary previous approaches in several key metrics.

\end{abstract}

 \section{Introduction}
 \label{sec:intro}

Among the most common ways humans learn to perform tasks is by
observing others performing the task and then modeling what they observed.  This is an incredibly
powerful ability; YouTube, for example, has how-to videos that include
demonstration on countless
topics.\footnote{ https://visme.co/blog/types-of-youtube-videos/,
https://medium.com/sponsokit/how-to-videos-are-the-most-popular-content-on-youtube-55103cf62232} However, while humans learn quite well from a small number of
demonstrations or a single demonstration, machine imitation-learning algorithms~\cite{behavioral_cloning_framework,gail,behavioral_cloning_from_observation}
typically require a large number of examples to begin to learn to
successfully perform a task. In many situations, large numbers of expert human demonstrations are not available to create accurate and performant imitators.

Another popular machine-learning approach for sequential tasks is Reinforcement Learning (RL). RL agents are not necessarily human-like, but are instead trained toward acting optimally in an environment given a reward structure. Indeed, RL has demonstrated remarkable success in
numerous domains including board games~\cite{alpha_zero1}, physical
control~\cite{ppo}, autonomous
driving~\cite{autonomous_driving_survey}, and video
games~\cite{openai_five,alpha_star}.  However, RL is not without its challenges. First, it requires a reward function, which measures the quality of every achieved state, but such a function might not always be available or easy to construct. 
Additionally, if a task is particularly complex, an RL agent might continuously explore without ever converging to its goal (the delayed-reward problem).
Finally, real-world examples used to train an RL algorithm are often limited and imperfect. The insufficient exposure to high-quality examples can result in a poor and biased performance.
As a result of these limitations, effective RL is frequently not practical.

This paper presents a robust imitation learning algorithm called Continuous Mean-Zero Disagreement-Regularized Imitation Learning (CMZ-DRIL).  CMZ-DRIL leverages ensemble-based uncertainty quantification to construct a training reward. The agent ensemble is trained on a set of expert trajectories.  CMZ-DRIL uses the ensemble uncertainty to construct a continuous reward function, achieving a mean of zero by subtracting the observed exponential average within the reward. This reward is optimized using RL.  In contrast to the continuous CMZ-DRIL reward, the original DRIL~\cite{dril} uses a clipped uncertainty cost that is a step function with 2 possible outputs.  We hypothesize that the smooth reward improves the RL's ability to find the disagreement-minimization direction. CMZ-DRIL exhibits enhanced task performance using limited expert demonstrations and bypasses the need to identify an appropriate value of a critical threshold variable for each environment. This advancement is highlighted by comparing the performance of CMZ-DRIL agents to that of agents trained with Behavioral Cloning (BC)~\cite{behavioral_cloning} and DRIL.

We evaluate CMZ-DRIL in three environments: two OpenAI Gym MuJoCo
environments, Half Cheetah and Hopper, and a simple path planning
environment, PyUXV~\cite{abc}. Experimental results indicate that CMZ-DRIL can significantly improve performance of the imitator compared to previous key imitation learning algorithms BC and DRIL.
CMZ-DRIL can close around half of the performance gap between BC
and training with the true environmental reward. CMZ-DRIL achieves this
performance by leveraging only expert trajectories and environmental interaction, and does not require
additional task information, like reward function, or environmental information.  One way CMZ-DRIL could be very useful is to bootstrap behavior for other learning approaches, like developmental Artificial Intelligence (AI), when a small number of demonstrations are available.

 \section{Related Work}
 \label{sec:related_work}
 Many algorithms have been proposed to address the issue of training a policy with low quality and/or low quantity expert data. The proposed algorithms vary in their approach but tend to encourage the imitator to stay close to states that have a sufficient number of expert demonstrations. By encouraging the imitator to stay near states with expert data, the imitator can more reliably predict the expert's actions and achieve better performance.

Deep Q-learning from Demonstrations (DQfD)~\cite{DQfD} and Deep Deterministic Policy Gradient from Demonstration (DDPGfD)~\cite{Vecerk2017LeveragingDF} are imitation-learning variants of Deep Q-Networks (DQN) that include demonstration data in the replay buffer and gives priority to demonstration data. The main criticism of these methods is that the demonstration data is underutilized.

Policy Optimization from Demonstration (POfD)~\cite{POfD} is a variant of the method Generative Adversarial Imitation Learning (GAIL)~\cite{gail} that reduces the sensitivity of GAIL to the quantity and quality of training data. GAIL uses a learned discriminator to distinguish between trajectories demonstrated by an expert and trajectories generated by the imitator. Since GAIL relies solely on the discriminator reward, training the discriminator with imperfect demonstrations can lead to poor performance. If only a small number of demonstrations are available, the discriminator may effectively ``memorize'' the demonstrations, preventing imitator progress. POfD claims to mitigate the issues of imperfect and underutilized demonstration data by optimizing the policy with respect to the discriminator output and environment reward. While POfD shows promising results, it relies on training an adversarial network in parallel with the policy which can make the algorithm unstable and difficult to tune. Additionally, POfD utilizes the environmental reward, which may not be available in certain environments.

Recently, Brantley \textit{et al.} introduced DRIL~\cite{dril}, whose approach is to first train an
ensemble of imitators on the available expert data using
BC~\cite{behavioral_cloning} and then train a second
agent using BC as well as RL with a reward that encourages the agent to stay in states with low disagreement between the ensemble agents.  The idea is that the agent ensemble will agree well in parts
of the state space where good expert data is available, so the new agent
will be trained to remain in areas where it can most confidently
model the expert data available.  DRIL can offer improved performance in many Atari environments when limited
demonstrations are available.

Like DRIL, CMZ-DRIL also quantifies uncertainty, or disagreement, as the
standard deviation between an ensemble of imitators trained on expert
data.  Unlike DRIL, CMZ-DRIL uses a continuous reward function that includes an exponential average of past disagreement so the mean reward is approximately zero and does not require the user to identify an environment-specific threshold that defines positive or negative disagreement.  DRIL rewards the agent either $+1$ or $-1$ if the ensemble
standard deviation is below or above a user-defined threshold
respectively. It is possible that this binary reward makes it difficult for the
learning agent to progress gradually toward behaviors with lower
disagreement in many settings.  This reward may also be most
frequently positive or most frequently negative for a given threshold,
training environment, and data set, which can incentivize the agent to
prolong episodes in potentially undesirable ways (if positive) or end
them quickly (if negative).

 \section{Approach}
 \label{sec:approach}
 CMZ-DRIL trains an agent using both BC and RL. For RL, CMZ-DRIL provides a continuous, mean-zero reward based on the standard deviation of an imitator ensemble.

In the course of developing CMZ-DRIL, a reward was tested that penalized proportionally to the ensemble's uncertainty level without providing any positive reward. While this reward improved the agent's performance in some environments, the negative value of the reward encouraged the agent to prematurely end episodes within the environment. For example, in tests performed during this study within the CartPole environment, the agent would quickly leave the environment's operational bounds to avoid accruing negative reward. To mitigate this unintended behavior, CMZ-DRIL was developed to provide a reward that averages to 0 over time.

CMZ-DRIL uses the reward $r_i=-\alpha (u_i - \bar{u}_i)$, where $u_i$ is the ensemble standard deviation at time step $i$, and $\bar{u}_i = \gamma \bar{u}_{i-1} + (1-\gamma) u_{i-1}$ is the exponential average of the uncertainty experienced by the agent. 
Here, $\alpha>0$ and $0<\gamma<1$ are constants. In experiments, the constants $\alpha = 10$, $\gamma = 0.99$, and $\bar{u}_{0}=0$ performed well. By using the standard deviation directly within the reward, uncertainty changes are gradual. We hypothesize that this semi-smooth reward can provide the learning agent a clearer sense of progress, thus allowing it to descend the uncertainty gradient. Additionally, by subtracting the exponential average within the reward calculation, CMZ-DRIL reduces the incentive to end the episode in a way that would cause the agent's performance to deviate from the expert's.

CMZ-DRIL trains an imitator agent in two steps: First, CMZ-DRIL uses the same expert data both to pretrain the imitator and to train the ensemble for uncertainty quantification. Then, CMZ-DRIL employs Proximal Policy Optimization (PPO) (\cite{ppo}) to train the imitator based on the ensemble standard deviation. After each PPO step, the imitator is trained for an additional epoch on the expert data, using the Negative Log Likelihood (NLL), so that the current policy continues to choose the same actions as the expert in regions of high expert-data concentration.

\begin{algorithm}
\caption{CMZ-DRIL}\label{alg:cap}
\begin{algorithmic}
\Require $\{O_e, A_e\}$, Expert observation-action pairs\\
\textbf{Pretrain:}\\
Train agent with $\{O_e, A_e\}$ using NLL\\
Train ensemble with $\{O_e, A_e\}$\\
\textbf{Train:}
\While{Policy is not converged}\\
Collect trajectory rollouts:\\
\hspace{5mm}  $r_i=-\alpha (u_i - \bar{u}_i))$\\
\hspace{5mm} $\bar{u}_{i+1} = \gamma \bar{u}_i + (1-\gamma) u_i$\\
Update agent with PPO\\
Train agent with $\{O_e, A_e\}$ using NLL
\EndWhile
\end{algorithmic}
\end{algorithm}

\begin{figure*}[ht]
\centering
\subfloat[Reward]{\includegraphics[scale=0.28]{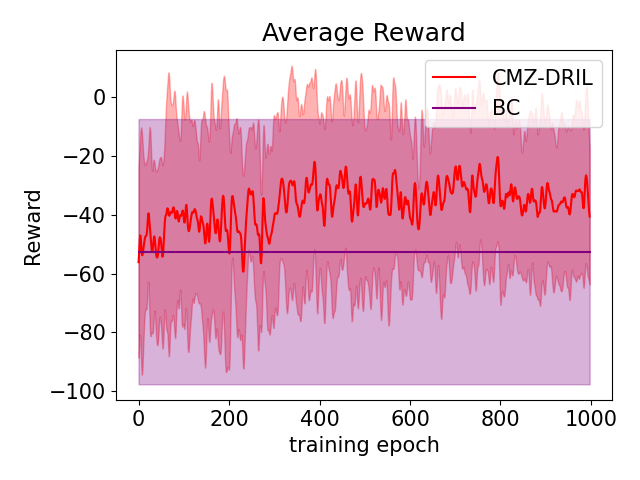}}
\subfloat[Frechet Distance]{\includegraphics[scale=0.28]{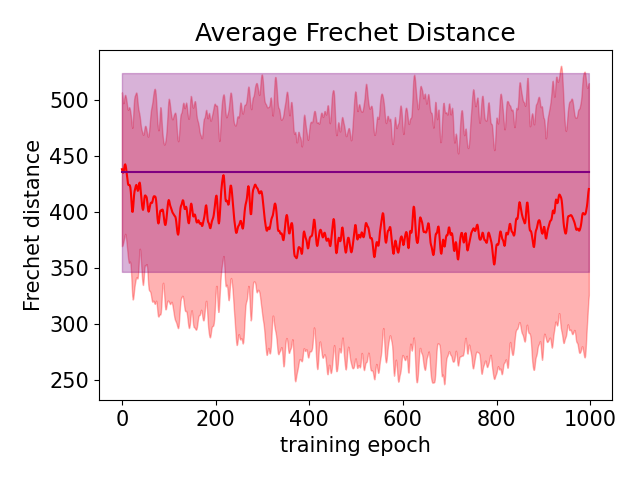}}
\subfloat[Action MSE]{\includegraphics[scale=0.28]{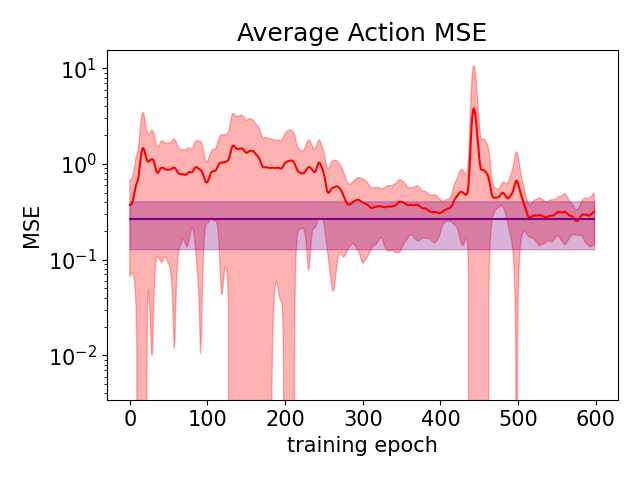}}\\

PyUXV

\subfloat[Reward]{\includegraphics[scale=0.28]{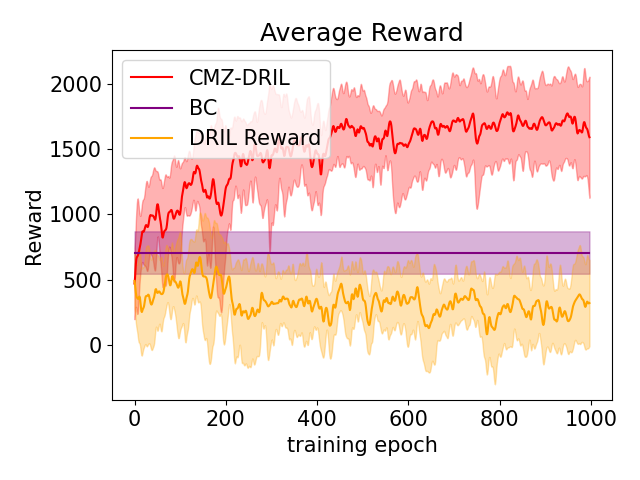}}
\subfloat[Frechet Distance]{\includegraphics[scale=0.28]{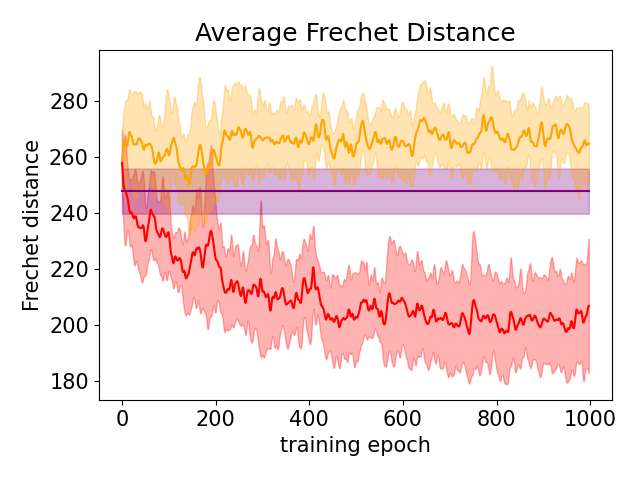}}
\subfloat[Action MSE]{\includegraphics[scale=0.28]{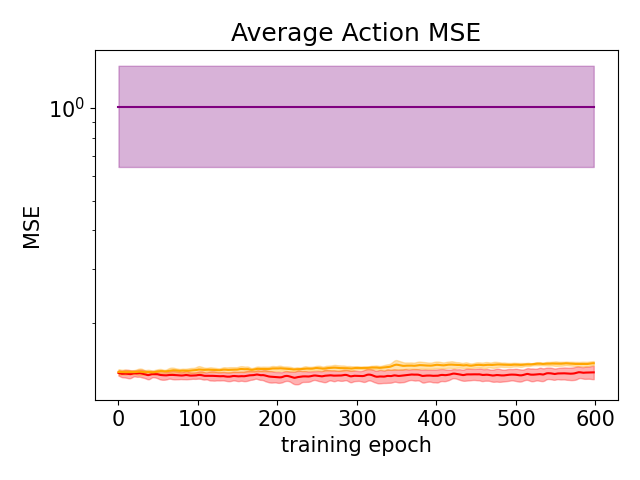}}\\
Half Cheetah

\subfloat[Reward]{\includegraphics[scale=0.28]{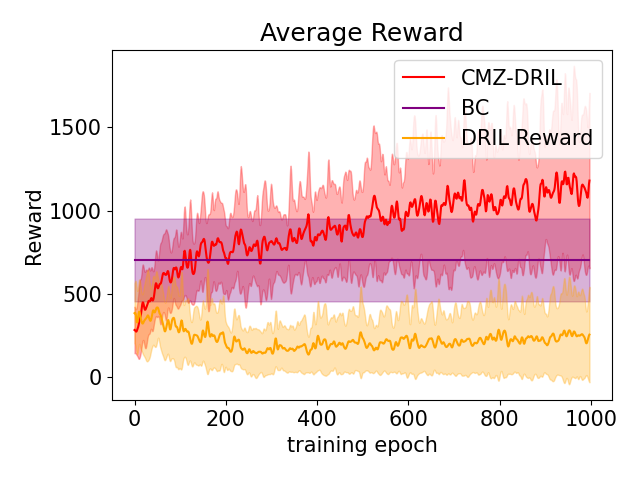}}
\subfloat[Frechet Distance]{\includegraphics[scale=0.28]{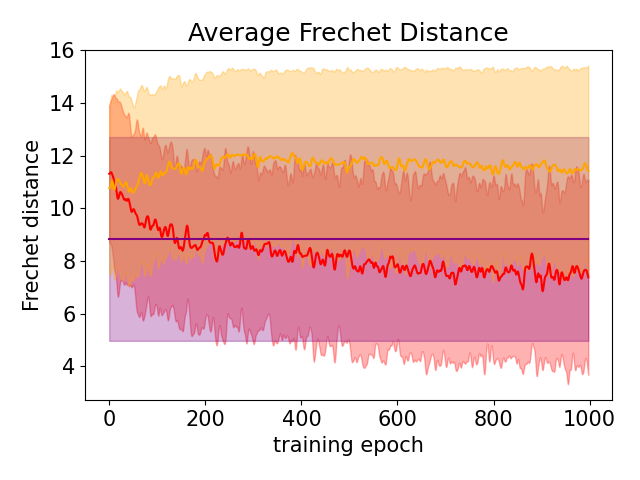}}
\subfloat[Action MSE]{\includegraphics[scale=0.28]{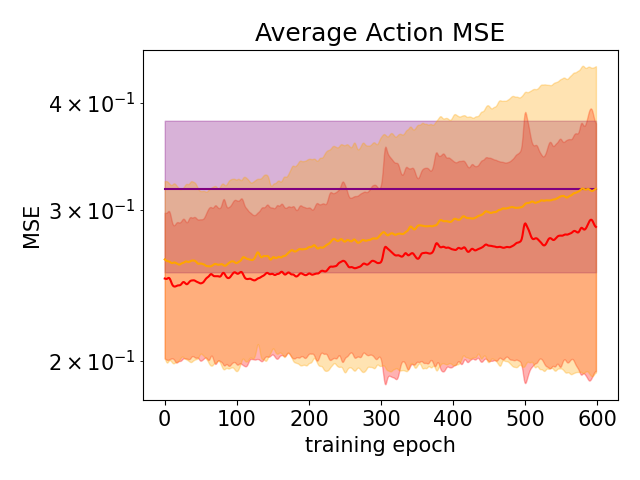}}\\
Hopper
\caption{Reward, Frechet Distance, and MSE comparisons between CMZ-DRIL, BC, and DRIL}
\label{fig:bc}
\end{figure*}

\begin{table*}[h]
    \centering
    \begin{tabular}{l|llllll}
             Method        & PyUXV     & Half Cheetah & Hopper\\\hline\hline
    Behavioral Cloning             & -52.6 (45.0) & 708 (162) &     704(248)                   \\
     DRIL Reward               & Not Tested  & 277 (60.9) &    244 (23.0)        \\
     CMZ-DRIL                 & -34.8 (29.4)  & 1690 (360) &    1140 (514)        \\
    \end{tabular}
    \caption{Mean reward (and standard deviation) for the last 100 training epochs of each environment.
    }
    \label{tab:eval_results}
\end{table*}

\begin{figure*}[h]
\centering
\subfloat[Reward]{\includegraphics[scale=0.28]{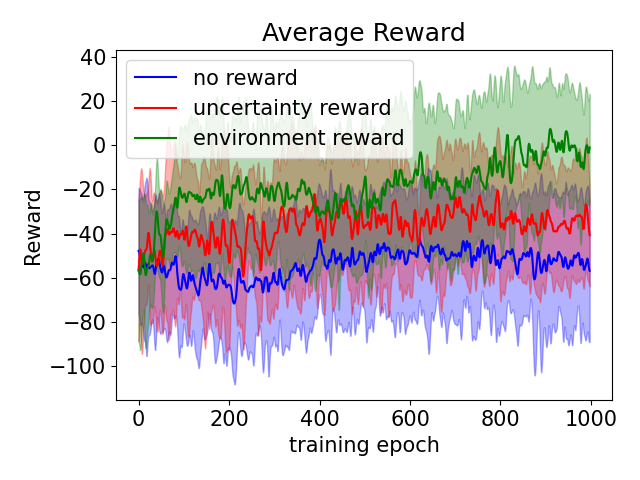}}
\subfloat[Frechet Distance]{\includegraphics[scale=0.28]{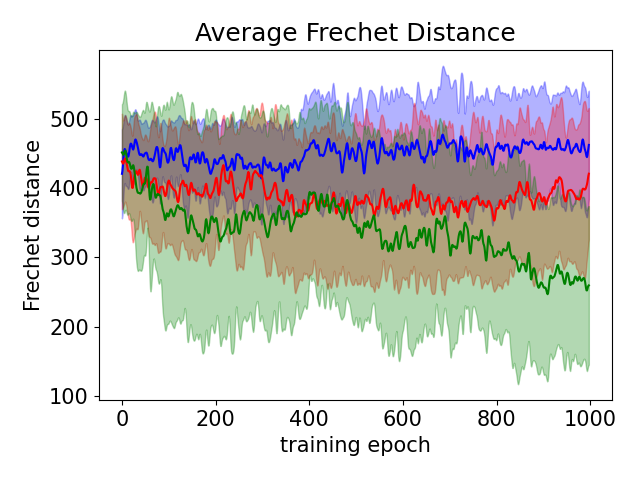}}
\subfloat[Action MSE]{\includegraphics[scale=0.28]{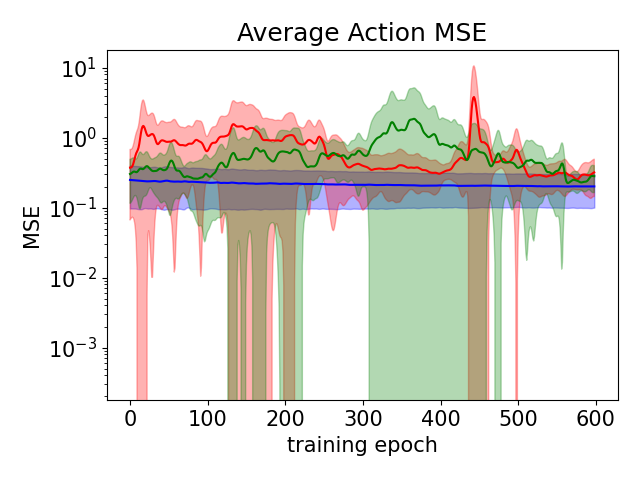}}\\
PyUXV

\subfloat[Reward]{\includegraphics[scale=0.28]{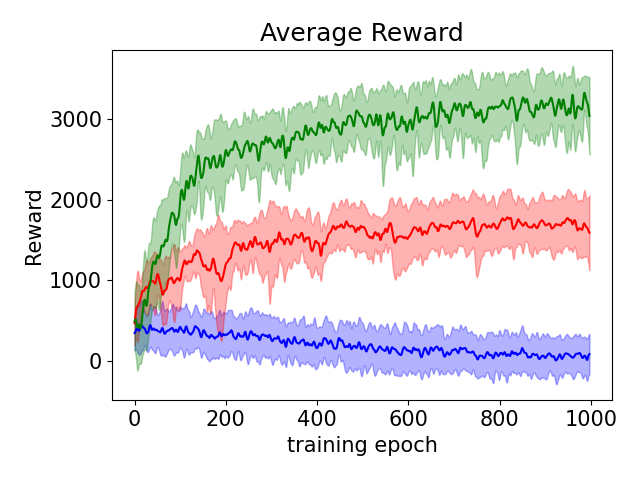}}
\subfloat[Frechet Distance]{\includegraphics[scale=0.28]{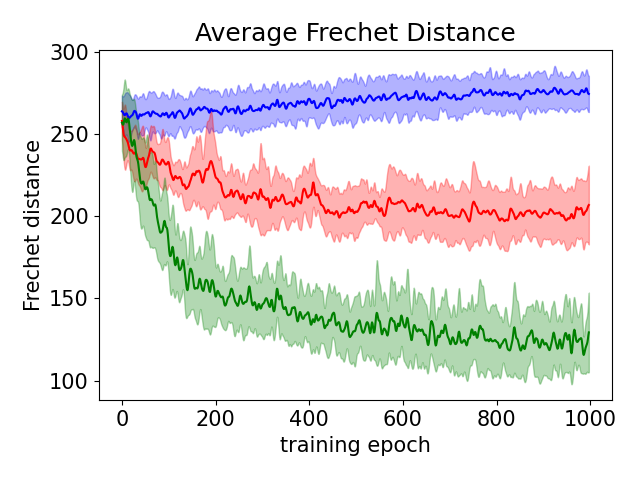}}
\subfloat[Action MSE]{\includegraphics[scale=0.28]{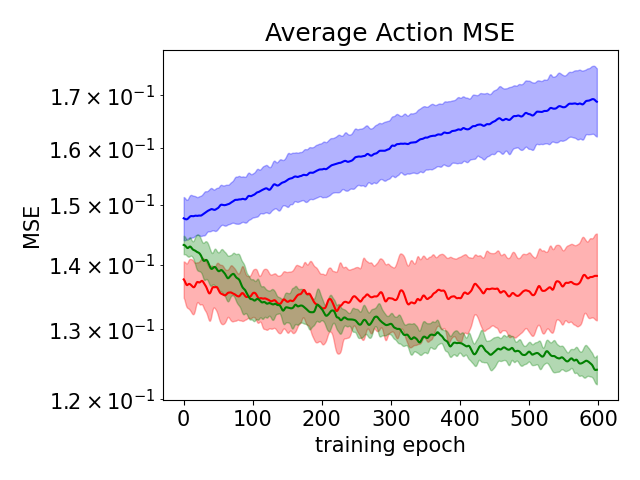}}\\
Half Cheetah

\subfloat[Reward]{\includegraphics[scale=0.28]{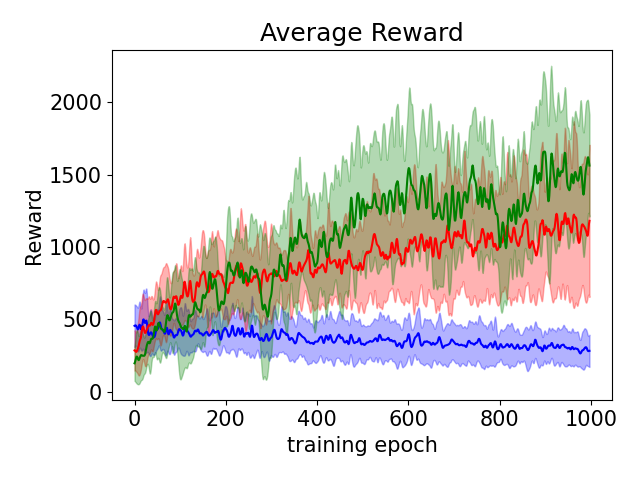}}
\subfloat[Frechet Distance]{\includegraphics[scale=0.28]{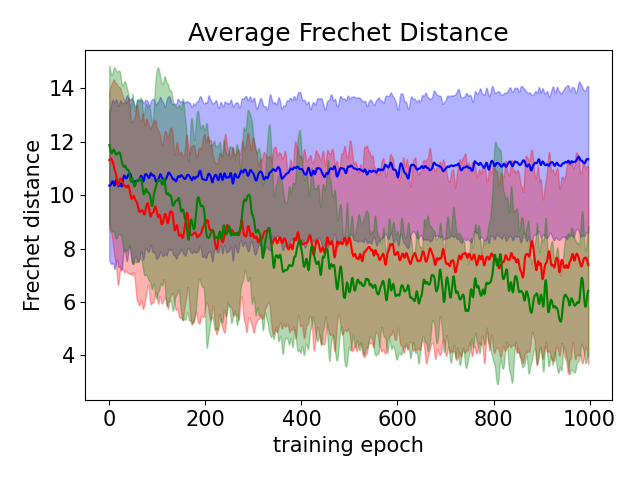}}
\subfloat[Action MSE]{\includegraphics[scale=0.28]{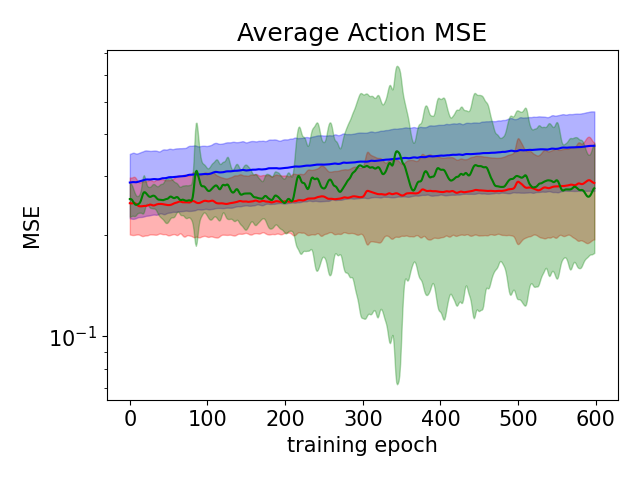}}\\
Hopper
\caption{Comparison of agent performance with CMZ-DRIL reward, true environment reward, and no reward}
\label{fig:reward_compare}
\end{figure*}

 \section{Experiments and Results}
 \label{sec:results}
 \subsection{Experiments}

Three environments, each with an expert agent, were used to test the performance of CMZ-DRIL.

\subsubsection{Environments}

\paragraph{PyUXV}
PyUXV is a waypoint-navigation and obstacle-avoidance environment. The observation space includes lidar readings as well as position and goal information (\cite{abc}). The agent uses continuous heading and throttle control to navigate in a 2D plane to the waypoint while avoiding a randomly-initialized field of obstacles. The expert for PyUXV is a reactive agent that minimizes its distance to the goal and avoids obstacles by following a tangent line to the obstacle.

\paragraph{MuJoCo Half Cheetah}
MuJoCo Half Cheetah (version 1) is a two-dimensional robot that controls a set of 8 joints to move itself forward. The goal in this environment is to run forward as fast as possible. The expert for Half Cheetah was generated using Soft Actor-Critic (SAC) from the implementation by \cite{stable-baselines3}.

\paragraph{MuJoCo Hopper}
MuJoCo Hopper (version 1) is a two-dimensional leg-like robot with controllable joints. The goal in this environment is to hop forward as fast as possible. The expert for Hopper was generated using SAC from the implementation by \cite{stable-baselines3}.

\subsubsection{Expert Demonstrations}

The set of expert demonstrations for one of our trials, i.e. the training data, includes 5 episodes for PyUXV, which consist of about 300 time-steps per training set,
and 1 episode for Half Cheetah and Hopper each with around 1000 steps.

\subsubsection{Trials}

We performed 5 training-and-evaluation trials on the MuJoCo environments for BC, CMZ-DRIL, and an implementation of DRIL that matches our implementation of CMZ-DRIL but with the DRIL reward and threshold parameters from the original DRIL paper~\cite{dril}.  We further compared CMZ-DRIL to BC on PyUXV although we did not include DRIL.  We did not have the DRIL threshold parameter for that environment.
Each trial used different training data, evaluation data, and network seeds. The same training and evaluation data were used for BC, DRIL, and CMZ-DRIL. We used 5 evaluation episodes for each experiment.

\subsection{Results}

Table~\ref{tab:eval_results} reports rewards achieved by BC, DRIL, and CMZ-DRIL.  Figure~\ref{fig:bc} reports training results. We smooth the epoch-series data using a Gaussian filter with $\sigma = 2$ to better visualize the pattern.

Through training an imitator to minimize uncertainty during an episode, CMZ-DRIL achieved significant improvement on environmental reward and a lower Frechet distance compared to BC. In contrast, trials using the DRIL reward did not exhibit significant improvement. Additionally, the original DRIL method did not show improvement in Half Cheetah and Hopper performance compared to BC within the HalfCheetahBulletEnv-v0 and HopperBulletEnv-v0 environments as reported by \cite{dril}.

Other than in the Half Cheetah environment, the Mean Squared Error (MSE) achieved by CMZ-DRIL is not significantly better than BC. In fact, the MSE does not appear to be correlated with the reward or Frechet distance metrics during the training process. It is possible that the reward for returning to known state space alters individual actions from what the expert would have done, leading to a higher MSE, even though the overall performance is more similar to the expert's.

Since CMZ-DRIL agents have improved performance compared to the BC agents, the contribution of ensemble-based reward to performance was also explored. Trials using the ensemble-based reward were compared to trials with no extra reward as well as trials in which the agent is trained with the true environmental reward. The trials that train with the true environmental reward represent an approximate upper bound on reward performance. Results are visualized in Figure \ref{fig:reward_compare}. The epoch-series data is smoothed using a Gaussian filter with $\sigma = 2$ to better visualize the pattern. CMZ-DRIL agents performed better in reward, Frechet distance, and MSE than those trained with no additional reward. As expected, the CMZ-DRIL agents performed worse than agents trained with the true environmental reward. The environmental reward encodes additional task information into the training process, and is able to achieve a higher performance compared to methods that only use expert data.
However, CMZ-DRIL was able to close the performance gap between agents trained with no reward and agents trained with the environmental reward by about $50\%$. The continuous, mean-zero reward provided by CMZ-DRIL reliably improves agent performance in the three environments tested.

 \section{Conclusions}
 \label{sec:conclusions}
 CMZ-DRIL can train uncertainty-aware imitators that noticeably outperform BC in a waypoint-navigation environment as well as in two MuJoCo environments with RL-based experts. Additionally, this paper demonstrates the improved performance of agents trained with the CMZ-DRIL reward as opposed to agents trained with no reward. CMZ-DRIL demonstrates the power of staying within the distribution of the training data for machine-learning agents. Imitation-learning through simple supervised learning can lead to imitators with paths that leave regions of high data concentration, putting the imitator in states where it has not learned to perform well. Methods that encourage agents to stay or return to regions of high data concentration have the potential to greatly improve performance and realism of the agents, particularly over longer time horizons.

CMZ-DRIL may be useful for bootstrapping learning agents for developmental AI when a small number of demonstrations is available.  It can give the agent some initial ability and direction to achieve its goals based on observations of other entities.  If coupled with an exploratory learning method, the CMZ-DRIL method could help the agent refine its performance as it slowly explores other regions of the statespace.

 \section*{Acknowledgments}
 This research is in part funded by the Test Resource Management Center (TRMC) and Test Evaluation/Science \& Technology (T\&E/S\&T) Program under contract W900KK-19-C-004.
Any opinions, findings, conclusions, or recommendations expressed in this material are those of the author(s) and do not necessarily reflect the views of the Test Resource Management Center (TRMC) or Test Evaluation/Science \& Technology (T\&E/S\&T) Program.

\bibliographystyle{unsrt}
 \bibliography{references}

\begin{thebibliography}{10}

\bibitem{behavioral_cloning_framework}
Michael Bain and Claude Sammut.
\newblock A framework for behavioural cloning.
\newblock {\em Machine Intelligence}, 103(15), 1995.

\bibitem{gail}
Jonathan Ho and Stefano Ermon.
\newblock Generative adversarial imitation learning.
\newblock In {\em Neural Information Processing Systems (NeurIPS)}, 2016.

\bibitem{behavioral_cloning_from_observation}
Faraz Torabi, Garrett Warnell, and Peter Stone.
\newblock Behavioral cloning from observation.
\newblock In {\em International Joint Conference on Artificial Intelligence
  (IJCAI)}, 2018.

\bibitem{alpha_zero1}
David Silver, Thomas Hubert, Julian Schrittwieser, Ioannis Antonoglou, Matthew
  Lai, Arthur Guez, Marc Lanctot, Laurent Sifre, Dharshan Kumaran, Thore
  Graepel, Timothy Lillicrap, Karen Simonyan, and Demis Hassabis.
\newblock A general reinforcement learning algorithm that masters chess, shogi,
  and {Go} through self-play.
\newblock {\em Science}, 362(6419):1140--1144, 2018.

\bibitem{ppo}
John Schulman, Filip Wolski, Prafulla Dhariwal, Alec Radford, and Oleg Klimov.
\newblock Proximal policy optimization algorithms, 2017.

\bibitem{autonomous_driving_survey}
B~Ravi Kiran, Ibrahim Sobh, Victor Talpaert, Patrick Mannion, Ahmad A.~Al
  Sallab, Senthil Yogamani, and Patrick Pérez.
\newblock Deep reinforcement learning for autonomous driving: {A} survey.
\newblock {\em {IEEE} Transactions on Intelligent Transportation Systems},
  23(6), 2022.

\bibitem{openai_five}
OpenAI, :, Christopher Berner, Greg Brockman, Brooke Chan, Vicki Cheung,
  Przemysław Dębiak, Christy Dennison, David Farhi, Quirin Fischer, Shariq
  Hashme, Chris Hesse, Rafal Józefowicz, Scott Gray, Catherine Olsson, Jakub
  Pachocki, Michael Petrov, Henrique~P. d.~O.~Pinto, Jonathan Raiman, Tim
  Salimans, Jeremy Schlatter, Jonas Schneider, Szymon Sidor, Ilya Sutskever,
  Jie Tang, Filip Wolski, and Susan Zhang.
\newblock {Dota} 2 with large scale deep reinforcement learning, 2019.
\newblock \url{https://arxiv.org/abs/1912.06680}.

\bibitem{alpha_star}
Oriol Vinyals, Igor Babuschkin, Wojciech~M. Czarnecki, Michaël Mathieu, Andrew
  Dudzik, Junyoung Chung, David~H. Choi, Richard Powell, Timo Ewalds, Petko
  Georgiev, Junhyuk Oh, Dan Horgan, Manuel Kroiss, Ivo Danihelka, Aja Huang,
  Laurent Sifre, Trevor Cai, John~P. Agapiou, Max Jaderberg, Alexander~S.
  Vezhnevets, Rémi Leblond, Tobias Pohlen, Valentin Dalibard, David Budden,
  Yury Sulsky, James Molloy, Tom~L. Paine, Caglar Gulcehre, Ziyu Wang, Tobias
  Pfaff, Yuhuai Wu, Roman Ring, Dani Yogatama, Dario Wünsch, Katrina McKinney,
  Oliver Smith, Tom Schaul, Timothy Lillicrap, Koray Kavukcuoglu, Demis
  Hassabis, Chris Apps, and David Silver.
\newblock Grandmaster level in {Starcraft II} using multi-agent reinforcement
  learning.
\newblock {\em Nature}, 575:350--354, 2019.

\bibitem{dril}
Kiant\'{e} Brantley, Mikael Henaff, and Wen Sun.
\newblock Disagreement-regularized imitation learning.
\newblock In {\em International Conference on Learning Representations (ICLR)},
  2020.

\bibitem{behavioral_cloning}
Michael Bain and Claude Sammut.
\newblock A framework for behavioural cloning.
\newblock In {\em Machine Intelligence 15}, 1995.

\bibitem{abc}
Corey~A. Lowman, Joshua~S. McClellan, and Galen~E. Mullins.
\newblock Imitation learning with approximated behavior cloning loss.
\newblock In {\em 2021 IEEE/RSJ International Conference on Intelligent Robots
  and Systems (IROS)}, pages 8921--8927, 2021.

\bibitem{DQfD}
Todd Hester, Matej Vecerik, Olivier Pietquin, Marc Lanctot, Tom Schaul, Bilal
  Piot, Dan Horgan, John Quan, Andrew Sendonaris, Gabriel Dulac-Arnold, Ian
  Osband, John Agapiou, Joel~Z. Leibo, and Audrunas Gruslys.
\newblock Deep q-learning from demonstrations, 2017.

\bibitem{Vecerk2017LeveragingDF}
Matej Vecer{\'i}k, Todd Hester, Jonathan Scholz, Fumin Wang, Olivier Pietquin,
  Bilal Piot, Nicolas Manfred~Otto Heess, Thomas Roth{\"o}rl, Thomas Lampe, and
  Martin~A. Riedmiller.
\newblock Leveraging demonstrations for deep reinforcement learning on robotics
  problems with sparse rewards.
\newblock {\em ArXiv}, abs/1707.08817, 2017.

\bibitem{POfD}
Bingyi Kang, Zequn Jie, and Jiashi Feng.
\newblock Policy optimization with demonstrations.
\newblock In Jennifer Dy and Andreas Krause, editors, {\em Proceedings of the
  35th International Conference on Machine Learning}, volume~80 of {\em
  Proceedings of Machine Learning Research}, pages 2469--2478. PMLR, 10--15 Jul
  2018.

\bibitem{stable-baselines3}
Antonin Raffin, Ashley Hill, Adam Gleave, Anssi Kanervisto, Maximilian
  Ernestus, and Noah Dormann.
\newblock Stable-baselines3: Reliable reinforcement learning implementations.
\newblock {\em Journal of Machine Learning Research}, 22(268):1--8, 2021.

\end{thebibliography}

\end{document}